\documentclass[a4,12pt]{article}

\usepackage[T1]{fontenc}        %pour avoir des accents dans le .dvi
\usepackage[latin1]{inputenc}   %pour taper des accents direct dans le texte
\usepackage{a4wide}             %marges normales

\usepackage{graphicx,amsmath}

\usepackage{natbib}

\title{Integration of navigation and action selection functionalities
  in a computational model of cortico-basal ganglia-thalamo-cortical
  loops}

\author{Benoît Girard$^{1,2,}$\footnote{\emph{correspondence to:}
    B. Girard, CNRS LPPA, Collège de France, 11 place Marcellin
    Berthelot, 75231 Paris Cedex 05, France, , \emph{Tel.:}
    +33-144271391, \emph{Fax:} +33-144271382 \emph{E-mail:}
    benoit.girard@college-de-france.fr} , David Filliat$^{3}$,
    Jean-Arcady Meyer$^{1}$, \\Alain Berthoz$^{2}$ and Agnès
    Guillot$^{1}$} 

\date{}
\begin{document}

%\twocolumn[%
\maketitle

% Ce qui suit : pour double interligne soumission AB
\begin{center}
%\vspace{1.5em}
$^1$ \emph{AnimatLab/LIP6, CNRS - University Paris 6} \\
$^2$ \emph{LPPA, CNRS - Collège de France} \\
$^3$ \emph{DGA/Centre Technique d'Arcueil} \\
\end{center}
%\newpage %SAB

%\begin{abstract}
{%\small % CAMREAD
This article describes a biomimetic control architecture affording an animat both action selection and navigation functionalities. It satisfies the survival constraint of an artificial metabolism and supports several complementary navigation strategies. It builds upon an action selection model based on the basal ganglia of the vertebrate brain, using two interconnected cortico-basal ganglia-thalamo-cortical loops: a ventral one concerned with appetitive actions and a dorsal one dedicated to consummatory actions.

The performances of the resulting model are evaluated in
simulation. The experiments assess the prolonged survival permitted by
the use of high level navigation strategies and the complementarity of
navigation strategies in dynamic environments. The correctness of the
behavioral choices in situations of antagonistic or synergetic
internal states are also tested. Finally, the modelling choices are
discussed with regard to their biomimetic plausibility, while the
experimental results are estimated in terms of animat adaptivity.
}
%\end{abstract}

\vspace{0.3cm}

Short title: Basal ganglia model of action selection and navigation.
\vspace{0.5cm}
%]

%\newpage %SAB
%\begin{multicols}{2}

%%%%%%%%%%%%%%%%%%%%%%%%%%%%%%%%%%%%%%%%
\section{Introduction}
%%%%%%%%%%%%%%%%%%%%%%%%%%%%%%%%%%%%%%%%%

The work described in this paper contributes to the \emph{Psikharpax}
project, which aims at building the control architecture of a robot
reproducing as accurately as possible the current knowledge of the
rat's nervous system (Filliat \emph{et al.}, 2004\nocite{filliat04}),
it thus concerns biomimetic modelling derived from data gathered with
rats. The main purpose of the \emph{Psikharpax} project is to refocus
on the seminal objective advocated by the animat approach: building "a
whole iguana" \citep{dennett78wh}, instead of designing
isolated and disembodied functions.  

Indeed, in the animat literature, a great deal of work is devoted to
the design of isolated control architectures that provide either action
selection or navigation abilities --two fundamental functions for an
autonomous system. The main objective of robotic navigation
architectures is to afford an animat with various orientation
strategies, like dead-reckoning, taxon navigation, place-recognition
or planning (Filliat and Meyer, 2003, Meyer and Filliat, 2003 for
reviews\nocite{filliat03map,meyer03map}).
The main objective of action selection architectures is to maintain
the animat into its ``viability zone'', defined by the state space of
its ``essential variables'' \citep{ashby52}, through efficient
switches between various actions (Prescott et al.,
1999\nocite{prescott99layer} for a review). Even if there is evidence
that an effective animat requires the use of these two
functionalities, few models attempt to integrate them, taking into
account the specific characteristics of each.  

On the one hand, most of the navigation models insert arbitration
mechanisms typical of action selection to solve spatial issues (e.g.,
Rosenblatt and Payton, 1989\nocite{rosenblatt89}), but they do not
take into account motivational constraints.

On the other hand, action selection models always integrate navigation
capacities ensuring an animat the ability to reach resources in the
environment, but they typically implement only rudimentary navigation
strategies --random walk and taxon navigation-- (e.g., Maes, 1991,
Seth, 1998\nocite{maes91,seth98evolv}).

The few models that process both navigation and action selection
issues are inspired by biological considerations, indicating that the
hippocampal formation, in association with the prefrontal cortex,
processes spatial information (O'Keefe and Nadel,
1978\nocite{okeefe78hippoccognitmap}), whereas the basal ganglia are
hypothesized to be a possible neural substrate for action selection in
the vertebrate brain \citep{redgrave99}. 

For example, Arleo and Gerstner (2000\nocite{arleo00spatial2}) propose
a model of the hippocampus that elaborates an internal map with the
creation of several ``place cells'', used by an animat to reach two
different kinds of resources providing rewards. 
The outputs of the model are assumed to be four action cells, coding
for displacements in cardinal directions, and assumed to belong to the
nucleus accumbens. This nucleus, located in the ventral part of the
basal ganglia, is hypothesized to integrate sensorimotor, motivational
and spatial information \citep{kelley99neural}. In this model, it
selects the actual displacement by averaging the ensemble activity of
the action cells. However, the animat does not select other navigation
strategies 
and does not have a virtual metabolism that puts constraints on the
timing and efficiency of the selection of its behaviors. 

Guazzelli et al. (1998\nocite{guazzelli98affor}) endow their simulated animat with two navigation strategies (place-recognition-triggered and taxon navigation, processed by hippocampus and prefrontal cortex) and homeostatic motivational systems (hunger and thirst, processed by hypothalamus). Here, the role of the basal ganglia is limited to computing of reinforcement signals associated with motivational states, while action selection properly occurs in the premotor cortex. Yet, in this work, there are no virtual metabolism constraints on action selection and because of the choice of a systems-interaction level of modelling, the internal operation of the modules is not specifically biomimetic.

Gaussier et al. (2000\nocite{gaussier00exper}) endow a motivated robot
(Koala\texttrademark, K-Team) with a virtual metabolism --generating
signals of hunger, thirst and fatigue-- and a topological navigation
capacity. A topological map is built in the hippocampus and used to
build a graph of transitions between places in the prefrontal cortex,
used for path planning.
The motor output is assumed to be effected by action neurons
in the nucleus accumbens, coding for three egocentric motions (turn
right, left, go straight). Motivational needs affect path planning by
spreading activation into the prefrontal graph from the desired
resources to the current location of the animat. They are transmitted
to the action neurons, allowing the animat to reach one goal by
several alternative paths, and to make compromises between different
needs. Here, one navigation strategy only is used, while various
complementary strategies coexist in animals.

These models do not entirely satisfiy the objectives of the fundamental functions, that is, dealing with survival constraints together with taking advantage of various complementary navigational strategies. Moreover, they do not exploit recent neurobiological findings concerning neural circuits devoted to the integration of these functions, involving two parallel and interconnected ``cortico-basal ganglia-thalamo-cortical'' loops (CBGTC, Alexander \emph{et al.}, 1986)\nocite{alexander86}, stacked on a dorsal to ventral axis, receiving sensorimotor (\emph{dorsal} loop) and spatial (\emph{ventral} loop) information. 

We previously tested a computational model of action selection,
inspired by the \emph{dorsal} loop and designed by Gurney \emph{et
  al.} (2001a,b\nocite{gurney01a,gurney01b}, referred to here as 'GPR'
after the authors'names), by replicating the \citet{montes-gonzalez00}
implementation in a survival task \citep{girard03}. To
improve the survival of an artificial system in a complex environment,
our objective is to add to this architecture a second circuit
--simulating the \emph{ventral} loop-- which selects locomotor actions
according to various navigation strategies: a \emph{taxon} strategy,
directing the animat towards the closest resource perceived, a
\emph{topological} navigation, building a map of the different
places in the environment and using it for path planning, together with
random \emph{exploration}, mandatory to map unknown areas and
allowing the discovery of resources by chance. The interconnection of the
\emph{dorsal} and \emph{ventral} loops is designed by means of
bioinspired hypotheses. The whole model will be validated in several
environments where the animat performs a simple survival task.  

After describing the navigation and action selection systems and how
they are interconnected, we will introduce the specific experimental
setup (survival task and animat configuration). The results will %first
concern tests on the animat's specific adaptive mechanisms and
behaviors, involving topological and taxon navigation, opportunistic
ability and conflict management in case of changes in the environment
or internal state. 

%%%%%%%%%%%%%%%%%%%%%%%%%%%%%%%%%%%%%%%%%%
\section{The control architecture}

This model has been introduced in a brief preliminary form in Girard \emph{et al.} (2004\nocite{girard04}).

\subsection{Navigation}\label{navigSys}

The choice of the navigation model was based on functional and efficiency criteria: it had to provide the animat with the capabilities of building a cognitive map, localizing itself with respect to it, storing the location of resources and computing directions to reach these resources; these operations had to be performed in real time and had to be robust enough to cope with the physical limitations of a real robot. The navigation system proposed by Filliat (\citeyear{filliat01cartog}) was chosen as it provides the required features and has been validated on a real robot (Pioneer\texttrademark, ActivMedia).

This model emulates hippocampal and prefrontal cortex functions. It builds a dense topological map in which nodes store the allothetic sensory input that the animat can perceive at respective places in the environment. These inputs are mean gray levels perceived by a panoramic camera in each of 36 surrounding directions, and sonar readings providing distances to obstacles in eight surrounding directions. A link between two nodes memorizes at which distance and in which direction the corresponding places are positioned relative to each other, as measured by the idiothetic sensors of odometry. The position of the animat is represented by a probability distribution over the nodes.

The model also provides an estimation of disorientation (\emph{D}), which varies from $0$ when the estimate of location is good, to $1$ when it is poor. $D$ increases when the robot is creating new nodes (it is in an unmapped area) and only decreases when it spends time in well known areas. The model also provides two $36$-component vectors indicating which directions to follow in order to either explore unmapped areas ($\mathbf{Expl}$) or go back to known areas in order to decrease disorientation ($\mathbf{BKA}$). If the animat does not regularly go back to known areas when it is very disoriented, the resulting cognitive map will not be reliable. Consequently, the addition of topological navigation to an action selection mechanism will put a new constraint on the latter, the one of keeping \emph{Disorientation} as low as possible.

We provided the model with the ability to learn the localization of resources important to survival (e.g. loading station, dangerous area) in the topological map. It is learned by associating active nodes of the graph with the type of resources encountered using Hebbian learning. By specifying the type of resource currently needed to a path planning algorithm applied on the graph, a vector $\mathbf{P}$ of $36$ values is produced, representing the proximity of that resource in $36$ directions spaced by 10°. Such a vector can be produced for each type of resource $res$, weighted by the motivation associated to that resource $m(res)$, and combined with the other ones to produce a generic path planning vector $\mathbf{Plan}$. The combination is processed as follows:

\begin{equation}\label{equ:fusionPlanifTerminale}
\mathbf{Plan} = 1 -\prod_{res}(1-m(res)\times\mathbf{P(}res\mathbf{)})
\end{equation}

%%%%%%%%%%%%%%%%%%%%%%%%%%%%%%%%%%%%%%%%%%
\subsection{Action Selection System}

--Figure \ref{FigGPR} around here--

The action selection model presented here is an extension of the one used in Girard \emph{et al.} (2003\nocite{girard03}), the GPR model (\nocite{gurney01a,gurney01b}Gurney \emph{et al.}, 2001a). It is a neural network model built with leaky-integrator neurons, in which each nucleus in the BG is subdivided into distinct channels each modelled by one neuron (Figure~\ref{FigGPR}), and each channel associated to an elementary action. Each channel of a given nucleus projects to a specific channel in the target nucleus, thereby preserving the channel structure from the input to the output of the BG circuit. The subthalamic nucleus (STN) is an exception as its excitation seems to be diffuse. 
Inputs to the BG channels are \emph{Salience} values, assumed to be computed in specific areas in the cortex, and representing the commitment to perform the associated action. They take into account internal and external perceptions, together with a positive feedback signal coming from the thalamo-cortical circuit, which introduces some persistence in the action performance. 
Two parallel \emph{selection} and \emph{control} circuits within the basal ganglia serve to modulate interactions between channels. Finally, the selection operates via disinhibition (Chevalier and Deniau, 1990\nocite{chevalier90disin}): at rest, the BG output nuclei are tonically active and keep their thalamic and motor system targets under constant inhibition. The output channel that is the less inhibited is selected, and the corresponding action executed.

A principal original feature of our model is that two parallel CBGTC loops are modelled, one selecting consummatory actions and the other appetitive actions.

\subsubsection{Dorsal loop}

In the BG, the \emph{dorsal loop} 
 implicated in the selection of motor responses in reaction to sensorimotor inputs and corresponds to the one modelled in the previous robotic studies of the GPR \citep{montes-gonzalez00,girard03}. 
Here we hypothesize that it will direct the selection of non-locomotor actions, which in the present case are limited to consummatory actions (robotic equivalents of eating, resting, etc.) (Figure~\ref{fullModel}). In this loop:

\begin{itemize}
\item input \emph{Saliences} are computed with internal and external sensory data;
\item at the output, a ``winner-takes-all'' selection occurs for the most disinhibited channel, as simultaneous partial execution of both reloading behaviors doesn't make sense.
\end{itemize}

--Figure \ref{fullModel} around here--

\subsubsection{Ventral loop}

The \emph{ventral loop} can be subdivided into two distinct subloops
\citep{thierry00hippoc}, originating from the core and shell regions
of its input nucleus (nucleus accumbens or NAcc) (Zham and Brog,
1992\nocite{zahm92commen}). In the present work, we will only retain
the core subloop (that will be henceforth also called \emph{ventral
  loop}), which has been proposed to play a role in navigation towards
rewarding places (Mulder \emph{et al.}, 2004; Martin and Ono, 2000\nocite{mulder04neuron,martin00effec}). 
The interactions between the hippocampus, the prefrontal cortex and
the NAcc core \citep{thierry00hippoc} could be the
substrate of a \emph{topological} navigation strategy. 
\emph{Taxon navigation} needs sensory information only and
could therefore be implemented in the \emph{dorsal} loop. However, it
was reported that the lesion of the NAcc also impairs object approach
\citep{seamans94selec}. This is why, in our model, this strategy will
also be managed by the \emph{ventral loop}.  

To summarize, we hypothesize that this loop will direct appetitive actions (robotics equivalent for looking for food, homing, etc.), suggesting displacements towards motivated goals (Figure~\ref{fullModel}). 

The \emph{ventral loop} is very similar --anatomically and
physiologically-- to the circuits of the \emph{dorsal} loop: the
dorsolateral ventral pallidum plays a role similar to the GP
\citep{maurice97posit}, the medial STN is dedicated to the ventral
circuits \citep{parent95funct2} as well as the dorsomedial part of the
SNr \citep{maurice99relat}. Thus, despite probable differences
concerning the influence of dopamine on ventral and dorsal input
nuclei, it is also designed by a GPR model. However, a few differences
are to be noted:
 
\begin{itemize}
\item \emph{Saliences} are computed with internal and external sensory
  data: the \emph{taxon} navigation needs distal sensory inputs to
  select a direction and all navigation strategies are modulated by
  the motivations. Additional data coming from the navigation system
  proposes motions on the basis of a \emph{topological} navigation
  strategy and map updates of current positions; 
\item each nucleus is composed of 36 channels, representing allocentric displacement directions separated by 10°;
\item the lateral inhibitions which occur in the nucleus accumbens core are no longer uniform as in the \emph{dorsal} loop, but increase with the angular distance between two channels (see eqn. \ref{latInhib}), so that close directions compete less than opposite ones;
\item at the output, the selection makes a compromise among all
  channels disinhibited above a fixed threshold. The direction chosen
  by the animat is computed by a vector sum of these channels,
  weighted by their magnitudes of disinhibition.  
\end{itemize}

\subsubsection{Interconnection of Basal Ganglia loops}\label{BGconnect}

Interconnections between the parallel CBGTC loops is needed to
  coordinate their respective selection processes. This is especially
  true here, when selections concerning navigation taken in the
  \emph{ventral} loop --like following a planned path leading to a
  resource-- might be conflicting with behavioral choices made by the
  \emph{dorsal} loop --like resting. Four main hypotheses concerning
  interconnections between loops have been proposed in the rat's
  brain. Two of them (\emph{Hierarchical pathway} \citep{joel94}
  and \emph{Dopaminergic hierarchical pathway}
  \citep{joel00connec}) were
  discarded because they only allow unidirectional communication from
  ventral to dorsal loops, whereas bidirectional or dorsal-to-ventral
  communication was necessary to solve our conflicts. The two
  remaining possibilities are (1) the \emph{Cortico-cortical pathway}:
  cortical interconnections between areas implied in different loops
  could allow bidirectional flows of information between loops;
  and (2) the \emph{Trans-subthalamic pathway}
  (\nocite{kolomiets01segreg,kolomiets03basal}Kolomiets et al., 2001,
  2003): the segregation of loops is not perfectly preserved at the
  level of the STN, some neurons belonging to one loop are excited by
  cortical areas belonging to other loops, thus, parts of the SNr
  belonging to one loop can be excited by another loop
  (Figure~\ref{fullModel}).

  We implemented the 
\emph{trans-subthalamic} hypothesis, by distributing \emph{dorsal} STN
activation to the \emph{ventral} outputs (see eqn. \ref{stnconnect}
and Figure~\ref{fullModel}). Selection of an action in the
\emph{dorsal} loop increases activity in the \emph{dorsal} STN, which
in turn increases activation of the \emph{ventral} outputs, preventing
any movement from occuring. 

The precise mathematical description of the resulting model is given
 in appendix \ref{ModeleGPRmath}.

\section{Experimental setup}

\subsection{Environment and survival task}\label{survive}

The experiments are performed in simulated 2D environments involving, as in Girard \emph{et al.} (\citeyear{girard03}), the presence of ``ingesting'' and ``digesting'' zones, but with the addition of ``dangerous'' places. The animat has to reach ``ingesting'' zones in order to acquire \emph{Potential Energy} ($E_P$), which it should convert into \emph{Energy} ($E$) in ``digesting'' zones, in order to use it for behavior. Note that a full load of $Energy$ allows the animat to survive only $33min$. Paths to reach these zones may contain dangerous areas to avoid.

The software used is a simulator programmed in C++, developed in our laboratory. Walls and obstacles are made of segments colored on a 256 level grayscale. The effects of lighting conditions are not simulated: the visual sensors have a direct access to the color. The three type of resources are represented by $50cm\times50cm$ squares of specific colors: the ``ingesting'' ($E_p$), ``digesting'' ($E$) and ``dangerous'' ($DA$) areas are respectively gray ($127$), white ($255$) and dark gray ($31$). They can be used by the animat when the distance between their centre and the centre of the animat is less than $70cm$ (i.e. when they occupy more than $60°$ of the visual field). The other gray objects have no impact on survival but help the navigation system discriminating places.

\subsection{The animat}

The animat is circular ($30cm$ diameter), and translation and rotation speeds are $40cm.s^{-1}$ and $10°.s^{-1}$ respectively. Its simulated sensors are: 
\begin{itemize}
\item an omnidirectional linear camera providing the color of the nearest segment for every $10°$ surrounding sector,
\item eight sonars with a $5m$ range, a directional incertitude of $\pm 5°$ and a $\pm 10cm$ distance accuracy,
\item encoders measuring self-displacements with an error of $\pm 5\%$ of the measured distance,
\item a compass with a $\pm 10°$ range of error of estimated direction.
\end{itemize}

The sonars are used by a low level obstacle avoidance reflex which overrides any decision taken by the BG model when the animat comes too close to obstacles. The navigation model uses the camera, encoders and compass inputs. The BG model uses the camera input to compute nine external variables: 
\begin{itemize}
\item Three $36$-component vectors, $\mathbf{Prox(}DA\mathbf{)}$, $\mathbf{Prox(}E_P\mathbf{)}$ and $\mathbf{Prox(}E\mathbf{)}$ providing the proximity of each type of resource in each direction. This measure is related to the angular size of the resource in the visual field with a 10° resolution, as it is obtained by counting the number of contiguous pixels of the resource color in a 7 pixels window centered on the direction considered. These vectors are the basis of the taxon navigation strategy.
\item Three variables, $mProx(DA)$, $mProx(E_P)$ and $mProx(E)$ which are the max values of the components of $Prox$ vectors.
\item Three Boolean variables, $A(DA)$, $A(E_P)$ and $A(E)$, which are true if the corresponding $mProx$ value is one (i.e. if the resource is less than $70cm$ away and thus usable).
\end{itemize}

These purely sensory inputs are completed by the vectors produced by the topological navigation system: the path planning vector $\mathbf{Plan}$, the exploration vector $\mathbf{Expl}$ and the ``go back to known areas'' vector $\mathbf{BKA}$.

The animat has four internal variables: \emph{Energy} and \emph{Potential Energy}, which concern the survival task  (see \ref{survive}), \emph{Fear}, which is a constant, fixing the strength of the repellent effect of ``dangerous areas'' and \emph{Disorientation}, which is provided by the topological navigation system (see \ref{navigSys}). From these variables are derived four motivations used in \emph{saliences} computations and in the weighting of the $\mathbf{Plan}$ vector (eqn. \ref{equ:fusionPlanifTerminale}). The motivations to go back to known areas and to flee dangerous areas are respectively equal to the \emph{Disorientation} and \emph{Fear} variables, while the motivation to reach \emph{Energy} and \emph{Potential Energy} resources are more complex:

\begin{equation}
\begin{array}{l}
m(DA) = F \\
m(BKA) = D \\
m(E) = (1 - E) \sqrt{1 - (1 - E_P)^2} \\
m(E_P) = 1 - E_P
\end{array}
\end{equation}

The variables used to compute saliences in each loop are summarized in
Figure~\ref{fullModel}, and the details of these computations are
given in appendix \ref{SalComput}.

\section{Experiments}

Three different experiments are carried out in simple environments in order to test the adaptive mechanisms the animat is provided with.

Experiment 1 tests the \emph{efficiency of the navigation/action selection models interface}. An animat capable of topological navigation has to survive in an environment containing one resource of \emph{Energy} and one resource of \emph{Potential Energy} which cannot be seen simultaneously. It is compared to an animat using the taxon strategy only, the use of the topological navigation is expected to improve the survival time. 

Experiment 2 tests adaptive action selection in a \emph{changing environment}: on the one hand, the animat has to use a taxon strategy in order to reach newly appeared resources; on the other hand, it has to forget the location of exhausted resources to head towards abundant ones.

Experiment 3 tests adaptive action selection in case of \emph{antagonistic or synergetic internal states}: on the one hand, in a situation where two paths lead to a resource and the shortest one includes a dangerous area; on the other hand, in a situation where a short path leads to one resource only, while a longer one leads to two resources satisfying two different needs.

In experiments 2 and 3, the animat is provided with a previously built
map of the environment in order to allow statistical comparison of
runs with identical initial conditions.

\subsection{Experiment 1: Efficiency of the navigation/action selection interface \label{expeTopo}}

In this experiment, an animat traverses the environment ($7m\times9m$) depicted in Figure~\ref{evt1}: it contains one resource of $E$ and one resource of $E_P$, but it is impossible to see one resource from the vicinity of the other. In the first model configuration (\emph{condition A}), the animat uses both object approach and topological navigation strategies, whereas in the other one (\emph{condition B}), the animat uses object approach only. The ``reactive'' animat (\emph{condition B}), following taxon strategy only, has to rely on random exploration to find hidden resources. In contrast, after a first phase of random exploration and map building, the animat in \emph{condition A} should be able to reach desired resources using its topological map.

--Figure \ref{evt1} around here--

Ten tests, with a four-hour duration limit, are run for both animats. \emph{Energy} and \emph{Potential Energy} are initially set to 1. The comparison of the median of survival durations for both sets shows that in \emph{condition A}, the animat is able to survive significantly longer ($p<0.01$, U-test, see Table~\ref{tabResexp1}) than the animat in \emph{condition B}.

--Table 1 around here--

In \citep{girard03}, action selection was only constrained by the virtual metabolism. Here, the addition of the \emph{topological navigation} system generates a new constraint of limiting \emph{Disorientation}. Yet it does not affect the efficiency of action selection, as the life span of animats is enhanced. 

\subsection{Experiment 2: Changing environment\label{expeChg}}

--Figure \ref{evt2} around here--

This experiment takes place in the $6m \times 6m$ environment depicted in Figure~\ref{evt2}, where the second \emph{Potential Energy} resource is not always present.

\subsubsection{New resources: Coordination of the navigation strategies\label{expeApparition}}

In this case, the second \emph{Potential Energy} resource is not present during the mapping phase, so that when the animat reaches the first intersection, it perceives a new resource that is unknown by the topological navigation system. The topological and the taxon strategies are thus competing, the first one suggesting to move to the distant resource ($E_P1$) and the second to the newly appeared and closer resource ($E_P2$). For all tests, the animat is initially placed on the same location shown in Figure~\ref{evt2} and lacks \emph{Potential Energy} ($E=1$ and $E_p=0.5$). The tests are stopped when the animat activates the $ReloadE_P$ action.

The control experiment consisting of ten tests in which resource $E_P2$ is not added, results in a repeatable behavior of the animat: it goes directly to $E_P1$ and activates the $ReloadE_P$ action when close enough to $E_P1$.
Three series of fifteen tests, with different weightings of the salience computations (variations of eqn.~\ref{SalComputVentral1} in appendix using the weights of Table~\ref{resNewRes}), are compared by counting how many times the animat chose one resource versus the other. The results are summarized in Table \ref{resNewRes}.

--Table 2 around here--

The first weighting corresponds to the configuration used in the previous experiment (eqn.~\ref{SalComputVentral1}). The path planning weight is larger than the taxon strategy one. As a result, the animat often ignores the new resource and chooses the memorized one. When the relative importance of the two strategies is modulated by progressively lowering the path planning weight, the behavior of the animat is modified and an opportunistic behavior, where it prefers the new and closest resource, can be obtained. 

Consequently, if our control architecture does not intrinsically
exhibit an opportunistic or a pure planning behavior, it can easily
be tuned to generate the desired balance between these two extremes.

\subsubsection{Exhausted resources: Forgetting mechanism}

In this situation, resource $E_P2$ is present during mapping but is
removed during the tests. The animat then has to ``forget'' its
existence in the map in order to go to the other resource.

Fifteen tests are carried out, with the animat initially placed on the same start location (see Figure~\ref{evt2}) lacking \emph{Potential Energy} ($E=1$ and $E_p=0.5$). The tests are stopped when the animat activates the $ReloadE_P$ action.

The animat first goes to the closest $E_P$ resource coded by the
topological navigation system: the near but absent $E_P2$
resource. The forgetting mechanism (implemented by the Hebbian rule
used to link resources with locations on the map) allows the animat to
finally leave this area and to reach resource $E_P1$. The
time necessary to forget $E_P2$ is estimated by subtracting the
duration of the most direct path leading from the start
position to $E_P1$ \emph{via} $E_P2$ ($46s$) to the duration of
each test. The mean duration is $178s$ ($\sigma=78$), i.e. $2$ minutes
and $58$ seconds (max value $5$ minutes). It is a bit long (almost
10\% of the $33$ minutes survival duration with a full charge of
$Energy$), but it can be reduced by simply modifying the gain of the
Hebbian rule.

This shows that the ability to forget, which is necessary to survive in environments where resources are exhaustible, operates correctly.

\subsection{Experiment 3: Antagonistic or synergetic internal states}

\subsubsection{Antagonistic internal states: Fear vs reloading need \label{expeDg}}

--Figure \ref{evt3} around here--

A first experiment is run in an environment ($10m \times 6m$) containing two $E_P$ resources and a dangerous area blocking direct access to the closest one (Figure~\ref{evt3}). The \emph{Dangerous Areas} affect the planning algorithm of the topological navigation system in an inhibitory manner. A path planning vector leading to dangerous areas is computed, multiplied by the level of \emph{Fear} and subtracted to the other planning vectors: the term $-m(DA) \times \mathbf{P(}DA\mathbf{)}$ is added to the computation of $\mathbf{Plan}$ described in eqn.~\ref{equ:fusionPlanifTerminale}.

The animat initially lacks \emph{Potential Energy} and its level of \emph{Fear} is fixed ($E=1$, $E_P<1$, $F=0.2$). When the \emph{Dangerous Area} is absent, the animat systematically chooses the closest resource ($E_P1$). However, when it is present, this inhibits the drive to go towards the $E_P1$ resource and the final choice of the $E_P$ resource should thus depend on the importance of the lack of energy.

--Table 3 around here--

Two series of 20 tests are carried out in order to induce conflicts between internal states depending on \emph{Fear} and $E_P$, respectively with a moderate ($E_P=0.5$) and a strong ($E_P= 0.1$) lack of $E_P$. 
As illustrated in Table~\ref{tab:choixExp2_2}, the inhibition generated by the \emph{Dangerous Area} in the first case is strong enough and the animat, despite the longer route, selects $E_P2$. In the second one, the need for \emph{Potential Energy} is stronger and the animat, despite the danger, selects $E_P1$. These two opposite tendencies are significantly different (Fischer's exact probability test, $p<0.01$).

This experiment shows that the animat may take risks in emergency situations and avoid them otherwise. But, more generally, it shows that it can exhibit, in an identical environmental configuration, different behavioral choices adapted to its conflicting internal needs, an essential property for a motivated animat.

\subsubsection{Synergetically interacting motivations}\label{expeGauss}

--Figure \ref{evtT} around here--

This task is inspired by a T-maze experiment proposed in Quoy \emph{et al.} (\citeyear{quoy02learn}) in order to study the behavior generated by the coupling of two motivations. The left branch of the T contains one $E_P$ resource while the right one contains both an $E$ and an $E_P$ resource (Figure~\ref{evtT}). The length of the right branch is varied so that the ratio of the right branch length to the left branch length is 1, 1.5 or 2. The animat is initially placed in the lower branch of the T, with a motivation for both $E$ and $E_P$ ($E=0.5$ and $E_P=0.5$). The test stops when the animat activates the $ReloadE_P$ action. In such a situation, the animat is expected to systematically prefer the right branch, even if it is longer, because choosing the left only satisfies the $E_P$ need, while choosing the right can satisfy both $E$ and $E_P$ needs.

--Table 4 around here--

Three series of fifteen tests are carried out with branch length ratio values of 1, 1.5 and 2, with an animat that needs both $E$ and $E_P$. As long as the ratio is not too high, the cumulated activation generated by the two resources on the right is higher than the drive generated by the single $E_P$ resource on the left (Table~\ref{tab:choixExpT}, ratio 1 and 1.5). However, when the two resources on the right are too far away, the drive they generate is attenuated by distance and the animat becomes more and more attracted by the resource on the left (Table~\ref{tab:choixExpT}, ratio 2).

The Gaussier \emph{et al.} (\citeyear{gaussier00exper}) model of
navigation integrates the notion of ``preferred path'' by reducing the
apparent distance between two nodes of the map when they are often
used. This allows the right branch to become preferred and thus
systematically chosen over time. Future development of our model
should include such a habit learning capability.

\section{Discussion}

The proposed biomimetic model integrates both navigation and action selection, in taking into account the specificities of both survival constraint and variety of navigation strategies. Simulations in benchmark environments validate 1) the survival advantage of using path planning strategies, 2) the benefits of simultaneously using taxon and planning strategies along with the necessity of being able to forget when operating in changing environments, and 3) the capability of the model to behave adaptively in case of conflicting and synergetic motivations.

\subsection{From Rattus rattus...}

How the brain coordinates the interface between spatial maps, motivation, action selection and motor control systems is of timely interest. The rat brain is widely investigated in this purpose, but many issues remain to be clarified. By synthesizing observed mechanisms in a behaving artificial system, our work helps to formulate several questions.

For example, our model points out limitations about the current
neurobiological knowledge concerning the actual role of NAcc core
channels: do they represent, as in our model and in e.g., Strösslin
(2004\nocite{strosslin04}), competing directions of movements? In
Experiment 2.1, the level of opportunism is fixed and does not adapt
to changing conditions (whereas taxon navigation is less reliable in
poor lighting conditions), as the \emph{ventral} loops selects one
direction taking into account all the navigation strategies. This
could be changed by having it selecting among the strategies the most
adapted one before a \emph{dorsal} loop selects the direction of
motion based on the chosen strategy suggestion only. Such coding has
recently received support by the work 
of Mulder \emph{et al.} (\citeyear{mulder04neuron}), on the basis of
electrophysiological recordings in hippocampal output structures
associated with the NAcc and a nucleus of the dorsal
stream (ventromedial caudate nucleus). Another and more complex role
may also be considered: NAcc core could interface goals, their
location, their amount and the corresponding motivations with
information coming from several neural structures like other limbic
structures or CBGTC loops \citep{dayan00rewar}. 

Likewise, our model questions the putative substrates of interactions
between CBGTC loops and their mode of operation, a subject of active
current research. We may have implemented the trans-subthalamic
hypothesis in an exaggerated manner. In fact the overlap of STN
projections from various loops is rather limited
\citep{kolomiets03basal}, while in our model they extensively reach
the whole output of the \emph{ventral loop}. This choice was indeed
convenient for the role attributed here to the dorsal and ventral
channels, respectively coding for immobile and mobile actions. Recent
results relative to interactions at the level of BG output projections
to dopaminergic nuclei in rats
\citep{mailly03threedimenorganrecuraxon} shed a new light on the
\emph{dopamine hierarchical pathway} and could be the basis of an
alternative model. In the GPR, varying the dopamine level affects
directly the ability to select, therefore, the possibility that one
loop may modulate the dopamine level of another one could be the basis
of an alternative mechanism for a loop to shunt another loop. One
cannot finally exclude the possibility that the resolution of
selection conflicts in the CBGTC loops is not only managed in the BG
but also in downstream brainstem structures, for example in the
reticular formation (Humphries \emph{et al.}, this issue).

\subsection{...to Psikharpax}

In Experiment 1, the planning animat (\emph{condition A})
sometimes dies because of a imperfect hand-tuning of the salience
computations, which causes it to stop to reload too far away from
resources. The basal ganglia, in interaction with the dopaminergic
system, is supposed to be the neural substrate for reinforcement
learning. In order to avoid such problems in the future, we are now
adding such a mechanism of automatic optimization of salience
computations to our model \citep{khamassi04comparcriticsr}. 

As mentioned in introduction, this work contributes to the
\emph{Psikharpax} project, which aims at building an artificial rat
(Filliat \emph{et al.}, 2004\nocite{filliat04}). As it evolves, this
artificial rat will be endowed with more than the few motivations
taken into account here, in the aim to improve the actual autonomy of
current robots, often devoted to a single task. The development of
polyvalent artifacts working in natural environments is indeed
promising for many applications in the home or in the office, as well
as future space programs with unmanned missions. Our work also helps
assessing the operational value of the biomimetic models used for this
purpose.

%%%%%%%%%%%%%%%%%%%%%%%%%%%%%%%%%%%%%%%%%%%%%
%\newpage %SAB
\appendix
\section{Appendix: Mathematical model description}

\subsection{GPR structure}\label{ModeleGPRmath}

Activation (a) of every neuron of the model:

\begin{equation}
\tau \frac{d a}{dt} = I - a
\end{equation}

where: I: input of the neuron, $\tau$: time constant ($\tau =
25ms$). Corresponding output (y):

\begin{equation}
y = 
\left\{
\begin{array}{rl}
0 & \mbox{if } a < \epsilon\\
m \times (a - \epsilon) & \mbox{if } \epsilon \le a < \epsilon + 1/m\\
1 & \mbox{if } \epsilon + 1/m \le a \\ 
\end{array}
\right.
\end{equation}

Values of $\epsilon$ and $m$ for each nucleus in Table~\ref{tab:paramsModeleAn}.

--Table 5 around here--

In each module (D1 and D2 striatum subparts, STN, EP/SNr,
GP, VL, TRN and cortical feedback), the input of each channel $i$ is
defined by the equations \ref{eqD1} to \ref{eqP}, where $N$: number of
channels, $S_i$: salience of channel i, $\lambda$: dopamine level ($0.2$).

\begin{equation}
I_{D1}^i = (1 + \lambda) S_i - \sum_{\substack{j=0 \\ j \neq i}}^{N} y_{D1}^i
\label{eqD1}
\end{equation}

\begin{equation}
I_{D2}^i = (1 - \lambda) S_i - \sum_{\substack{j=0 \\ j \neq i}}^{N} y_{D2}^i
\label{eqD2}
\end{equation}

In our model of the \emph{ventral} loop, lateral inhibitions (sum terms in eqn. \ref{eqD1} and \ref{eqD2}) increase with the angular difference between two channels. They are replaced in the \emph{ventral} loop by the following $LI$ term:

\begin{equation}
LI^i = \sum_{\substack{j=0 \\ j \neq i}}^{N} \frac{|i-j| mod (N/2)}{N/2} \times y_{(D1 \mbox{ or } D2)}^i
\label{latInhib}
\end{equation}

\begin{equation}
I_{STN}^i = S_i - y_{GP}^i
\end{equation}

\begin{equation}
I_{EP}^i = - y_{D1}^i - 0.4 \; y_{GP}^i + 0.8 \sum_{j=0}^{N} y_{STN}^j
\end{equation}

The \emph{trans-subthalamic} pathway is modelled by a modified
input for the \emph{ventral} EP/SNr (v and d stand for
\emph{ventral} and \emph{dorsal}): 

\begin{equation}
\begin{split}
I_{EPv}^i = & - y_{D1v}^i - 0.4 \; y_{GPv}^i \\
            & + 0.8 \sum_{j=0}^{N} y_{STNv}^j + 0.4 \sum_{j=0}^{N} y_{STNd}^j \\
\end{split}
\label{stnconnect}
\end{equation}

\begin{equation}
I_{GP}^i =  - y_{D2}^i + 0.8 \sum_{j=0}^{N} y_{STN}^j
\end{equation}

\begin{equation}
I_{VL}^i = y_{P}^i - y_{EP}^i - 0.13 \sum_{\substack{j=0 \\ j \neq i}}^{N} y_{TRN}^j
\end{equation}

\begin{equation}
I_{TRN}^i = y_{VL}^i + y_{P}^i
\end{equation}

\begin{equation}
I_{P}^i = y_{VL}^i
\label{eqP}
\end{equation}

%%%%%%%%%%%%%%%%%%%%%%%%%%%%%%%%%%%%%%%%%%%%%%%%%
\subsection{Salience computations}\label{SalComput}

The modification to the GPR model proposed in Girard \emph{et al.} (\citeyear{girard03}) consisted in allowing, for the computation of saliences, the use of sigma-pi neurons and non-linear transfert function applied to the inputs. This was kept in the present model and is the origin of the square roots and multiplications in the following equations.

%%%%%%%%%%%%%%%%%%%%%%%%%%%%%%%%%%%%%%%%%%%%%%%%%
\subsubsection{Experiments 1 and 2}

\emph{Dorsal} loop saliences ($E$ and $E_P$ reloading actions):

\begin{equation}
\begin{split}
  S_E = & \; 0.4 \times P_{E} + 1.2 \times A(E) \times m(E) \\
        & + 0.6 \times mProx(E) \times m(E) \\
\end{split}
\label{salE1}
\end{equation}

\begin{equation}
\begin{split}
  S_{E_P} = & \; 0.4 \times P_{E_P} + A(E_P) \times m(E_P) \\
                & + 0.2 \times mProx(E_P) \times m(E_P) \\
\end{split}
\label{salEp1}
\end{equation}

\emph{Ventral} loop salience for each direction i:

\begin{equation}
\label{SalComputVentral1}
\begin{split}
             S_i = & \; 0.2 \times P_{i} + W_{plan} \sqrt{\mathbf{Plan}_i}\\
                   & + 0.55 \sqrt{\mathbf{Prox(E)}_i} \times m(E)\\
                   & + W^{E_p}_{taxon} \sqrt{\mathbf{Prox(E_P)}_i} \times m(E_P)\\
                   & + 0.4 \times \mathbf{BKA}_i \times m(BKA)\\
                   & + \mathbf{Exp}_i \times (0.25 \\
                   & \quad + 0.05 \times (1-mProx(E_P)) \times m(E_P)\\
                   & \quad + 0.05 \times (1-mProx(E)) \times m(E))\\
\end{split}
\end{equation}

Where $W_{plan}$ and $W^{E_p}_{taxon}$ are respectively set to $0.65$ and $0.55$, except in experiment 4.2.1, where they take the values recorded in Table~\ref{resNewRes}.

%%%%%%%%%%%%%%%%%%%%%%%%%%%%%%%%%%%%%%%%%%%%%%%%%
\subsubsection{Experiment 3.1}

Saliences of the \emph{dorsal} loop computed as in experiments 1 and
2. \emph{Ventral} saliences modified to include the avoidance of
dangerous areas: 

\begin{equation}
\label{SalComputVentral2}
\begin{split}
             S_i = & \; 0.2 \times P_{i} + 0.45 \sqrt{\mathbf{Plan}_i}\\
                   & + 0.35 \sqrt{\mathbf{Prox(E)}_i} \times m(E)\\
                   & + 0.35 \sqrt{\mathbf{Prox(E_P)}_i} \times m(E_P)\\
	           & + 0.19 \times (1-\mathbf{Prox(DA)}_i) \times m(DA)\\
                   & + 0.4 \times \mathbf{BKA}_i \times m(BKA)\\
                   & + \mathbf{Exp}_i \times (0.05 \\
                   & \quad + 0.05 \times (1-mProx(E_P)) \times m(E_P)\\
                   & \quad + 0.05 \times (1-mProx(E)) \times m(E))\\
\end{split}
\end{equation}

%%%%%%%%%%%%%%%%%%%%%%%%%%%%%%%%%%%%%%%%%%%%%%%%%
\subsubsection{Experiment 3.2}

Experiment 3.2 showed that the weight of the \emph{dorsal} computations had to be lowered:

\begin{equation}
\begin{split}
  S_E = & \; 0.4 \times P_{E} + 0.9 \times A(E) \times m(E) \\
        & + 0.1 \times mProx(E) \times m(E) \\
\end{split}
\label{salE2}
\end{equation}

\begin{equation}
\begin{split}
  S_{E_P} = & \; 0.4 \times P_{E_P} + 0.9 \times A(E_P) \times m(E_P) \\
                & + 0.1 \times mProx(E_P) \times m(E_P) \\
\end{split}
\label{salEp2}
\end{equation}

The \emph{ventral} salience computations from experiments 1 and 2
risked stopping the animat too far from resources. As this problem
arose systematically in experiment 3.2, the term $(0.65
\sqrt{\mathbf{Plan}_i})$ term was changed for $(0.55
\sqrt{\mathbf{Plan}_i} \times (1-mProx(E)) \times (1-mProx(E_P))$. 

%\newpage %SAB

\section*{Acknowledgments}

We thank Sidney Wiener for valuable discussions and proofreading of the manuscript. This research has been funded by the LIP6 and the Project \emph{Robotics and Artificial Entities} (ROBEA) of the French Centre National de la Recherche Scientifique.

\bibliographystyle{apalike}
\bibliography{BG.bib}

%%%%%%%%%%%%%%%%%%%%%%%%%%%%%%%%%%%%%%%%%%%%%%%%%%%%%%%%%%%%%
\newpage 

\begin{table}[h]
\caption{
  Comparison (U-test) of experiments testing median survival duration of animats in conditions A (taxon navigation only) and B (taxon and topological navigation).
  \vspace*{5pt}}

  \begin{center}
  \begin{tabular}{lrl}
    \hline
	{\bfseries Durations (s)} & {\bfseries Median} & {\bfseries Range} \\
    \hline
    \bfseries{A} & 14431.5 & 2531 : 17274\\
    \bfseries{B} & 4908.0  & 2518 : 8831\\
    \hline
    \bfseries{U test} & U = 15 & $p<0.01$ \\
    \hline
  \end{tabular}\label{tabResexp1}
  \end{center}

\vspace*{-13pt}
\end{table}

\newpage 

\begin{table}[h]
  \caption{\label{resNewRes} Resource choice depending on the relative weighting of the two navigation strategies in the salience computation. $W_{plan}$ and $W^{E_p}_{taxon}$: weights related to planning and taxon navigation strategies respectively (see eqn.~\ref{SalComputVentral1}).}
  \begin{center}
    \begin{tabular}{ccrr}
      \hline
        \multicolumn{2}{c}{\bfseries Weights} & \multicolumn{2}{c}{\bfseries Choices} \\
      \hline
        $\mathbf{W_{plan}}$ & $\mathbf{W^{E_p}_{taxon}}$ & $\mathbf{E_P1}$ & $\mathbf{E_P2}$\\

      \hline
        0.65 & 0.55 & 13 & 2 \\
        0.55 & 0.55 & 7 & 8 \\
        0.45 & 0.55 & 2 & 13 \\
      \hline
    \end{tabular}
  \end{center}
\end{table}

\newpage 

\begin{table}[h]
  \caption{\label{tab:choixExp2_2}
    Resource choice depending on the initial $E_P$ level.
  }  
  \begin{center}
    \begin{tabular}{ccrr}
      \hline
        \multicolumn{2}{c}{\bfseries{Internal}} & \multicolumn{2}{c}{\bfseries{Incidence of}} \\
        \multicolumn{2}{c}{\bfseries{state}} & \multicolumn{2}{c}{\bfseries{choices}} \\
      \hline
        $\mathbf{F}$ & $\mathbf{E_{P}}$ & $\mathbf{E_P1}$ & $\mathbf{E_P2}$\\
      \hline
        0.2 & 0.1 & 13 & 7 \\
        0.2 & 0.5 & 2 & 18 \\
      \hline
        \multicolumn{2}{c}{\bfseries{Fisher's test}} & \bfseries{p<} & 0.01 \\ 
      \hline
    \end{tabular}
  \end{center}
\end{table}

\newpage 

\begin{table}[h]
  \caption{\label{tab:choixExpT}
    Branch choices depending on the length ratio.
  }  
  \begin{center}
    \begin{tabular}{crr}
      \hline
        \bfseries{ } & \multicolumn{2}{c}{\bfseries{Incidence of}} \\
        \bfseries{ } & \multicolumn{2}{c}{\bfseries{first choice}} \\
      \hline
        \bfseries{Ratio} & \bfseries{Left} & \bfseries{Right} \\
      \hline
        \bfseries{1} & 3 & 12 \\
        \bfseries{1.5} & 4 & 11 \\
        \bfseries{2} & 8 & 7 \\
      \hline
    \end{tabular}
  \end{center}
\end{table}

\newpage 

\begin{table}[h]
\caption{\label{tab:paramsModeleAn} Parameters of the transfer functions of the GPR model.}
\begin{center}
\begin{tabular}{lrr}
\hline
\bfseries GPR Module & $\mathbf{\epsilon}$ & $\mathbf{m}$ \\
\hline
D1 Striatum & 0.2   & 1    \\
D2 Striatum & 0.2   & 1    \\
STN         & -0.25 & 1    \\
GP          & -0.2  & 1    \\
EP/SNr      & -0.2  & 1    \\
Ctx         & 0     & 1    \\
TRN         & 0     & 0.5  \\
VL          & -0.8  & 0.62 \\
\hline
\end{tabular}
\end{center}
\end{table}

%%%%%%%%%%%%%%%%%%%%%%%%%%%%%%%%%%%%%%%%%%%%%%%%%%%%%%%%%%%%%%%%%%%%%%%%%%%%%%%%%%%%%%%%%%%%%%%%%%
% FIGURES :

\newpage 

\section*{Figure Captions}

\begin{figure*}[h]
\centerline{\includegraphics[width=\linewidth]{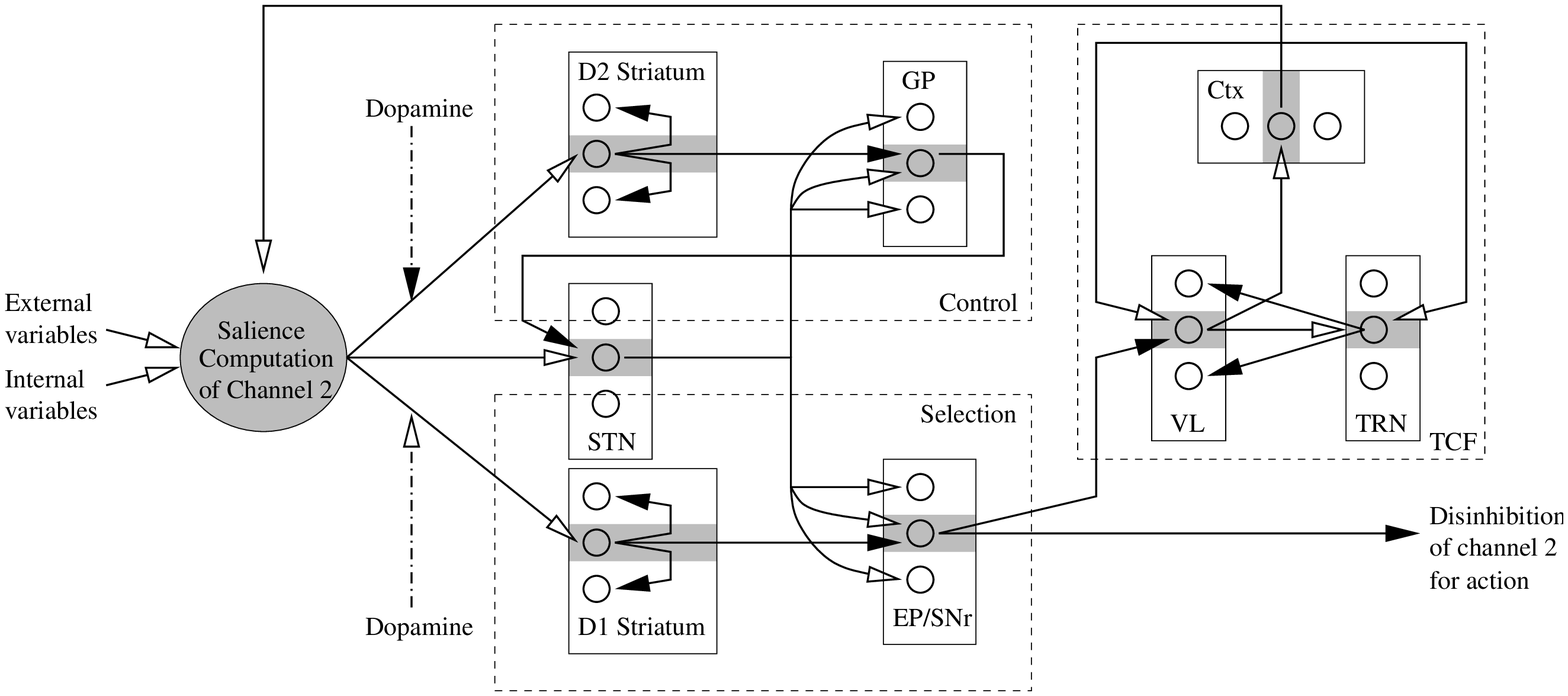}}
\caption{\label{FigGPR}}
\end{figure*}

\begin{figure*}[h]
\centerline{\includegraphics[width=\linewidth]{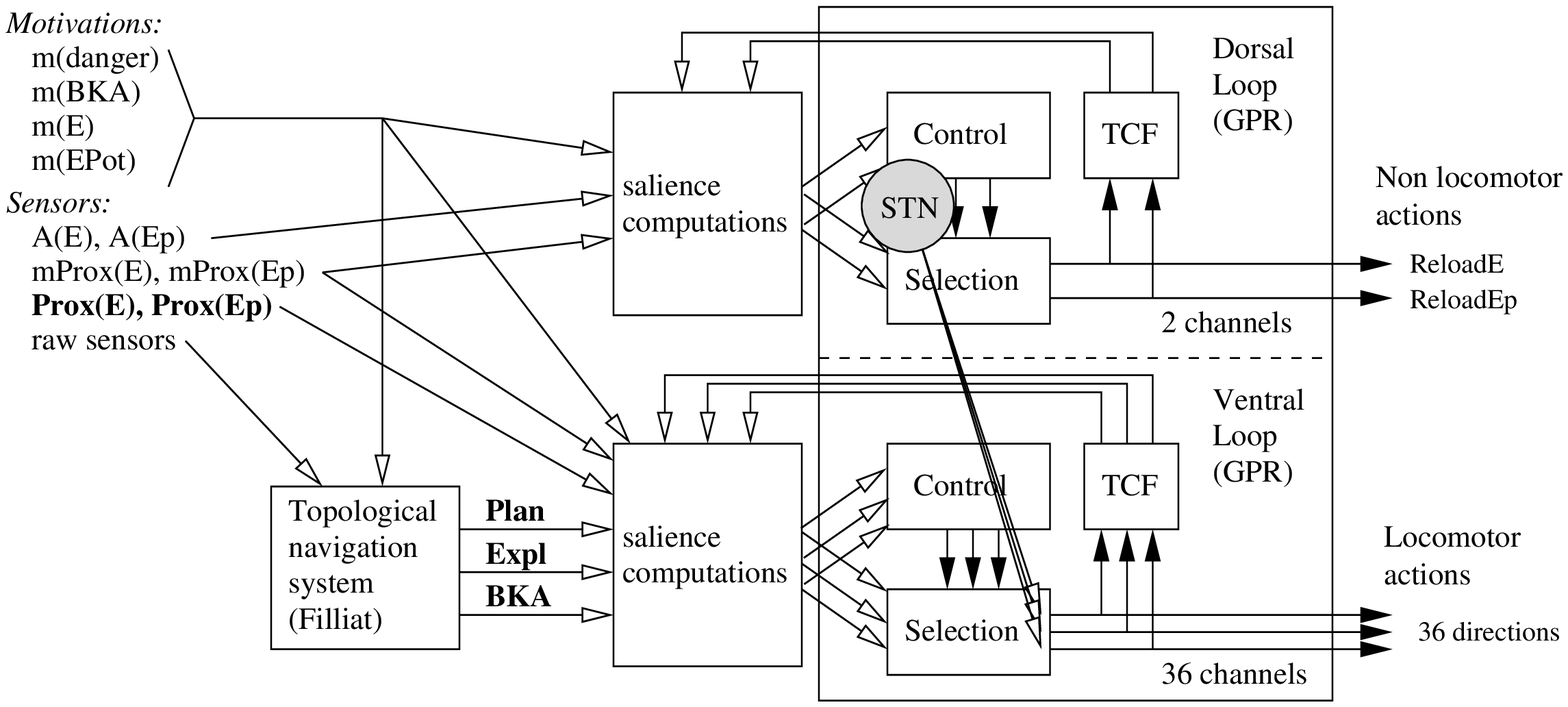}}
\caption{\label{fullModel}}
\end{figure*}

Figure \ref{FigGPR}: The GPR model. Nuclei are represented by boxes, each circle in these nuclei represents an artificial leaky-integrator neuron. On this diagram, three channels are competing for selection, represented by the three neurons in each nucleus. The second channel is represented by gray shading. For clarity, the projections from the second channel neurons only are represented, they are similar for the other channels. White arrowheads represent excitations and black arrowheads, inhibitions. D1 and D2: neurons of the striatum with two respective types of dopamine receptors; STN: subthalamic nucleus; GP: globus pallidus; EP/SNr: entopedoncular nucleus and substantia nigra pars reticulata; VL: ventrolateral thalamus; TRN: thalamic reticular nucleus. Dashed boxes represent the three subdivisions of the model proposed by its authors (Selection, Control of selection and thalamo-cortical feedback or TCF), note that these subdivisions appear on the simplified sketch of Figure~\ref{fullModel}.

Figure \ref{fullModel}: Final model structure. Input variables are exhaustively listed, 36-component vectors are in bold type. The excitatory projections from the STN of the dorsal loop to the EP/SNr of the ventral loop, which are the substrate for loops coordination, are highlighted.

\begin{figure}[h]
\centerline{\includegraphics[width=0.8\linewidth]{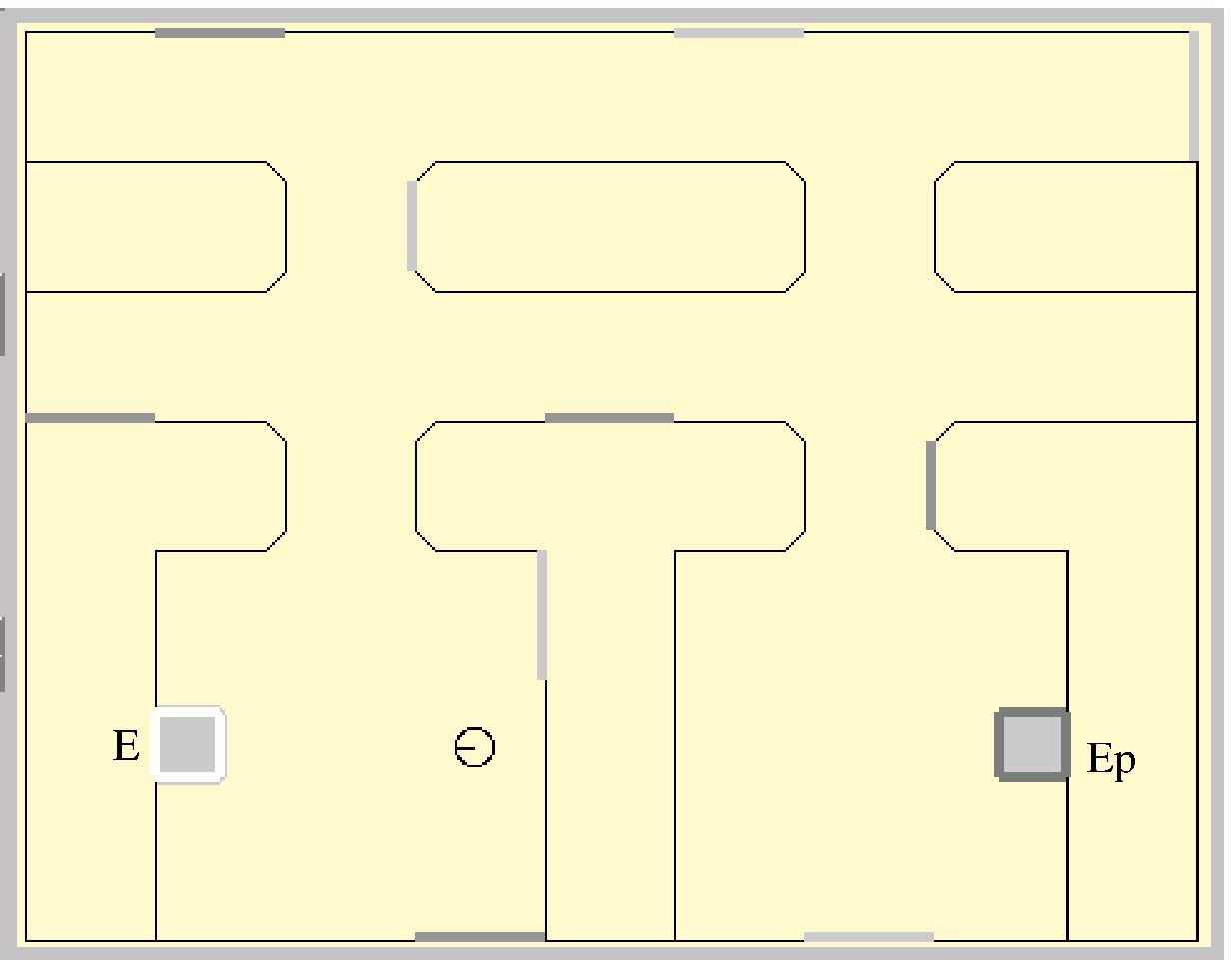}}
\caption{\label{evt1}}
\end{figure}

Figure \ref{evt1}: Experiment 1 environment. Initial position and orientation are represented by the schematic animat. $E$: \emph{Energy} resource; $E_P$: \emph{Potential Energy} resource. 

\begin{figure}[h]
\centerline{\includegraphics[width=0.6\linewidth]{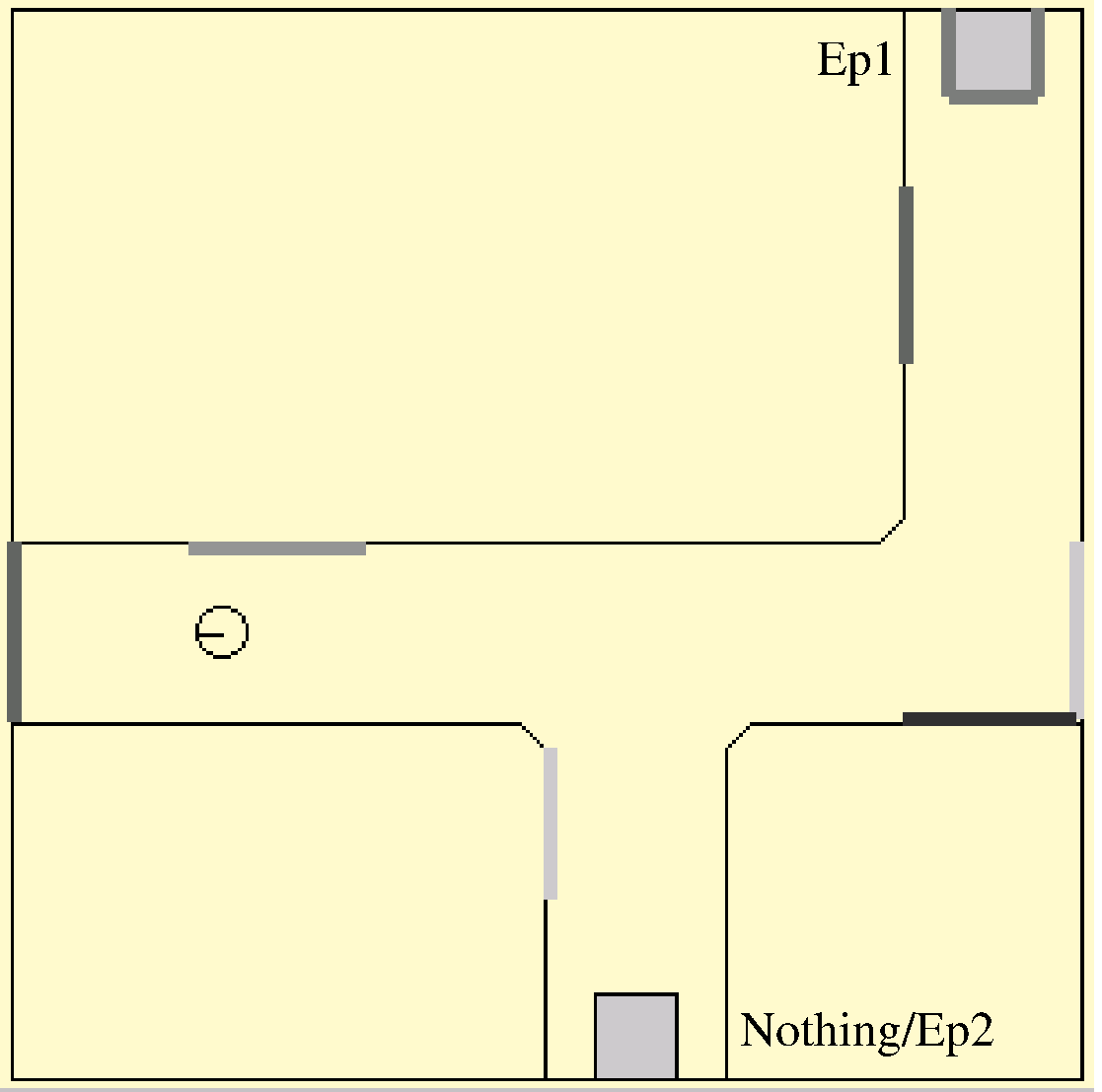}}
\caption{\label{evt2}}
\end{figure}

Figure \ref{evt2}: Experiment 2 environment. Initial position and orientation are represented by the schematic animat. $E_P$: \emph{Potential Energy} resource; $E_P2$ is absent in some experiments, see text. 

\begin{figure}[h]
\centerline{\includegraphics[width=0.8\linewidth]{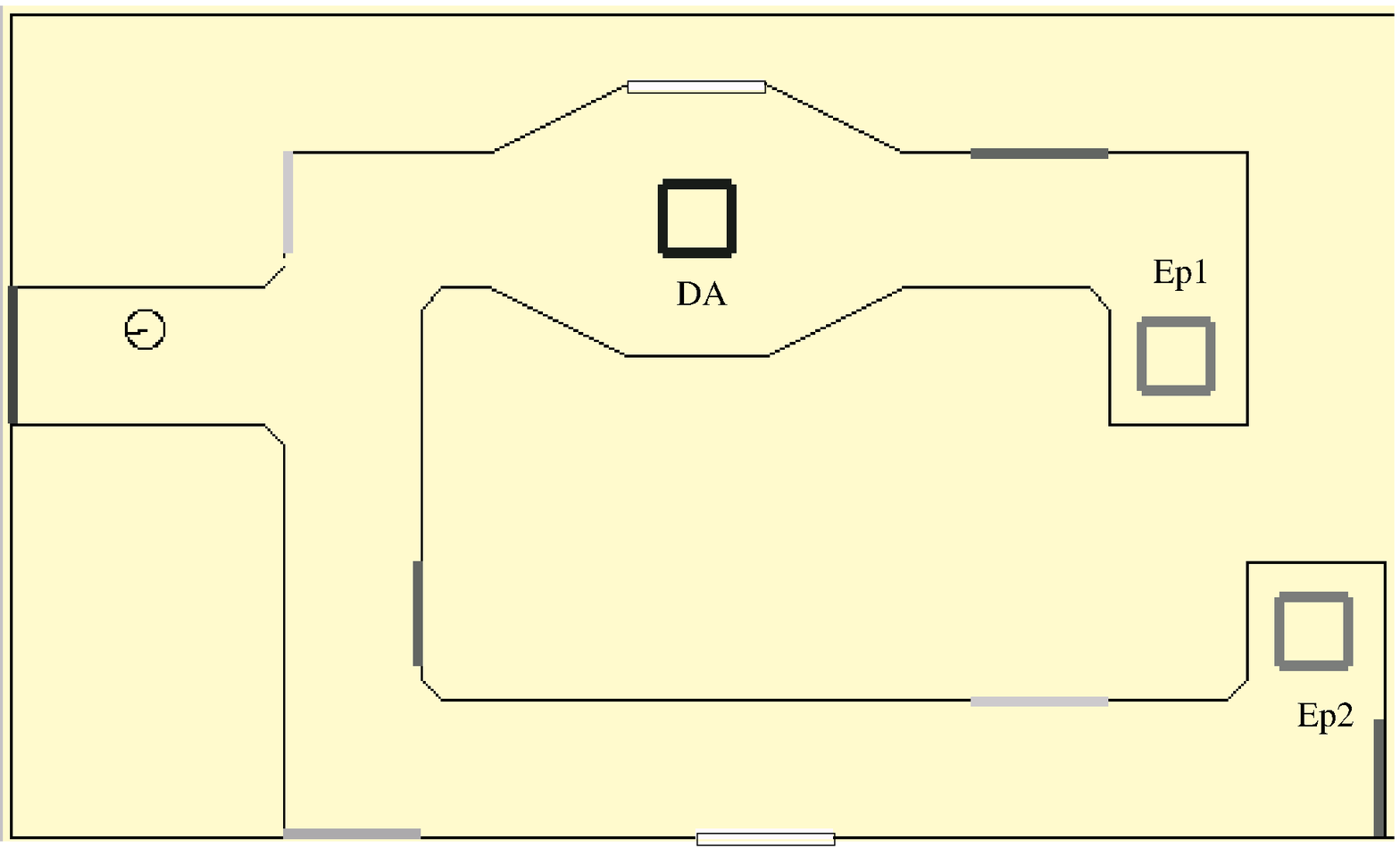}}
\caption{\label{evt3}}
\end{figure}

Figure \ref{evt3}: Experiment 3 environment. Initial position and orientation are represented by the schematic animat. $E_P1$,$2$: \emph{Potential Energy} resources; $DA$: dangerous area. 

\begin{figure}[h]
\centerline{\includegraphics[width=\linewidth]{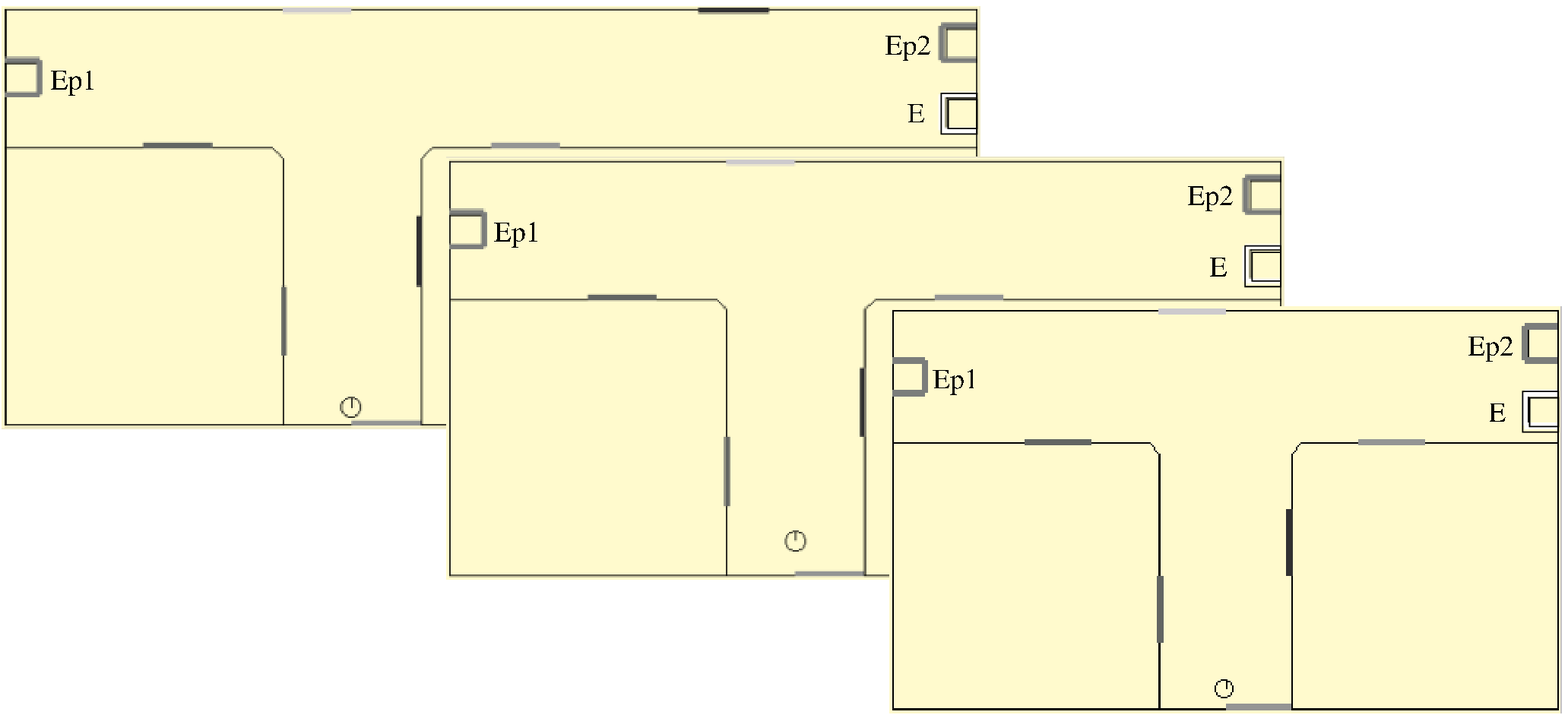}}
\caption{\label{evtT}}
\end{figure}

Figure \ref{evtT}: The three environments of experiment 4. The ratio of the right branch length to the left branch length varied between 1 and 2. Initial position and orientation is represented by the schematic animat. $E_P1$,$2$: \emph{Potential Energy} resources; $E$: \emph{Energy} resource. 

\newpage 

\newpage 

\newpage 

\newpage 

\newpage 

\newpage 

\end{document}